# Question-Answering Based Summarization of Electronic Health Records using Retrieval Augmented Generation

Walid Saba[1], Suzanne Wendelken[2] and James. Shanahan[3]

[1] Institute for Experiential AI, Northeastern University, 100 Fore St, Portland, ME 04101 USA
[2] The Roux Institute, Northeastern University, 100 Fore St, Portland, ME 04101 USA
[3] Institute for Experiential AI, Northeastern University, 100 Fore St, Portland, ME 04101 USA

**Abstract**

Summarization of electronic health records (EHRs) can substantially minimize 'screen time' for both patients as well as medical personnel. In recent years summarization of EHRs have employed machine learning pipelines using state of the art neural models. However, these models have produced less than adequate results that are attributed to the difficulty of obtaining sufficient annotated data for training. Moreover, the requirement to consider the entire content of an EHR in summarization has resulted in poor performance due to the fact that attention mechanisms in modern large language models (LLMs) adds a quadratic complexity in terms of the size of the input. We propose here a method that mitigates these shortcomings by combining semantic search, retrieval augmented generation (RAG) and question-answering using the latest LLMs. In our approach *summarization is the extraction of answers to specific questions that are deemed important by subject-matter experts (SMEs)*. Our approach is quite efficient; requires minimal to no training; does not suffer from the 'hallucination' problem of LLMs; and it ensures **diversity**, since the summary will not have repeated content but diverse answers to specific questions.

## 1. Introduction

Clinicians spend a considerable amount of time summarizing vast amounts of textual information that documents, for example, a patient's history, admission or discharge notes, administrated procedures, and recommended medications [6, 7]. Observational studies show that physicians spend 2-3 times the amount of time spent during patient encounters for chart review (33%), documentation (24%), and ordering (17%) (see [9]), amounting to approximately 3-5 hours per day of the physicians' time spent interacting with the EHR. By automatically summarizing all this textual data, keeping important and critical information while ignoring spurious details, minimizes 'screen time' and gives clinicians more time to spend with their patients [1] (see figure 1). The advent of large language models (LLMs) and their utility in a number of natural language processing (NLP) tasks have prompted many to employ these models in generating summaries for electronic health records (EHRs) but, apparently, results have been less than satisfactory. As reported in [6], these approaches suffer from the difficulty of adapting an LLM to the domain and where fine tuning does not seem to remedy this problem completely. The problem of available training data and the shortcomings of LLMs in handling domain-specific data has also been reported in [2]. Another issue with these approaches is lack of 'contextual' summarization since different users might require different types of information extracted from the EHRs. In our question-based approach, the content of the summary extracted is exactly the kind of content that ***answers specific questions*** posed by some subject-matter expert (SME). Moreover, our approach is a zero-shot learning approach that does not suffer from the bottleneck of the time-consuming effort of preparing a training dataset. Finally, our approach can be applied to any domain where contextualized summaries are required by different users, and where different users are interested in different questions being answered.

**Figure 1**: Left- the "anatomy" of a patient chart which includes sequential structured data such as lab results and vital signs interspersed with unstructured data such as clinical notes and radiological reports. Right - An example chart summation task to summarize the relevant information in a lengthy discharge note.

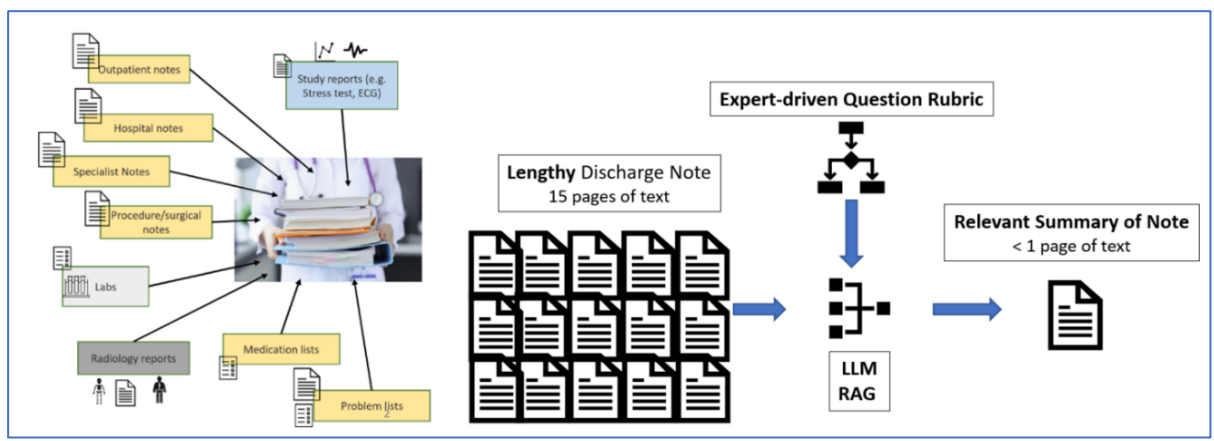

## 2. Methods

### 2.1 Data Source

The MIMIC-III dataset is a freely available public database containing de-identified EHR data, including clinical notes, lab results, and radiology reports, from 46,147 patients who were admitted to critical care units in the Beth Israel Deaconess Medical center between 2001 and 2012. We found this dataset to be the most comprehensive and representative dataset for a patient's EHR, specifically because it contains sequential clinical notes. In total, there are more than 2 million clinical notes contained in this dataset. We will also explore the use of public derivative datasets containing annotated notes (e.g., [10, 11]).

### 2.2 Dataset Creation and Quality Assurance

Patients with "complete" EHR data for a hospital admission period, as determined by the presence of an admission note, multiple clinical notes, and a discharge note will be selected for analysis. We will then down select at random 50 patients with long hospital stays (>10 days). Data will be inspected for distribution of patient demographics. Three clinician subject-matter-experts, including Co-Pi Wendelken, annotate discharge notes and provide target expert summaries.

### 2.3 Summarization by RAG-powered Question-Answering

Our basic strategy is to first segment EHR data into paragraphs, since paragraphs are the smallest unit of text that are semantically complete (see [5]). As shown in figure 2 (a), paragraphs are then indexed by their embeddings in a vector database[2].

    Summarization proceeds by looping through our relevant questions (whose answers are deemed by subject-matter experts to constitute a good summary)[3]. For each question we execute a query to our vector database to retrieve relevant snippets of text from paragraph segmented EHRs. Using the retrieved text we prepare a context to the prompt that is then executed on the LLM (the new prompt is

---

[2] For now we are using the Chroma vector database for indexing embeddings and k-nearest neighbor semantic search (https://www.trychroma.com/) but we eventually intend to use AWS RAG and Question-Answering Framework (see this)

[3] In [4] a somewhat similar question-based summarization approach is suggested, although the questions are assumed to be extracted from the text itself and thus the quality of the summary will depend on the quality of the questions extracted. Moreover, the approach is quite inefficient since the generation of good questions from the raw text is a computationally a very intensive process.

the initial question + retrieved content). The LLM will return an answer as well as the sentence where the answer was found, along with a confidence score about the answer. At the same time, the embeddings of noun phrases extracted from both question and answers are also matched (using cosine similarity). Each answer then has two scores, and a weighted average score is computed as the final score. Answers (sentences) with a score above some threshold are considered to be good answers/sentences and are collected to be considered sentences that should be part of the final summary. The final set of candidate sentences that have all the answers are then processed for some linguistic (resolve references, translate acronyms to the more meaningful synonyms, etc.)

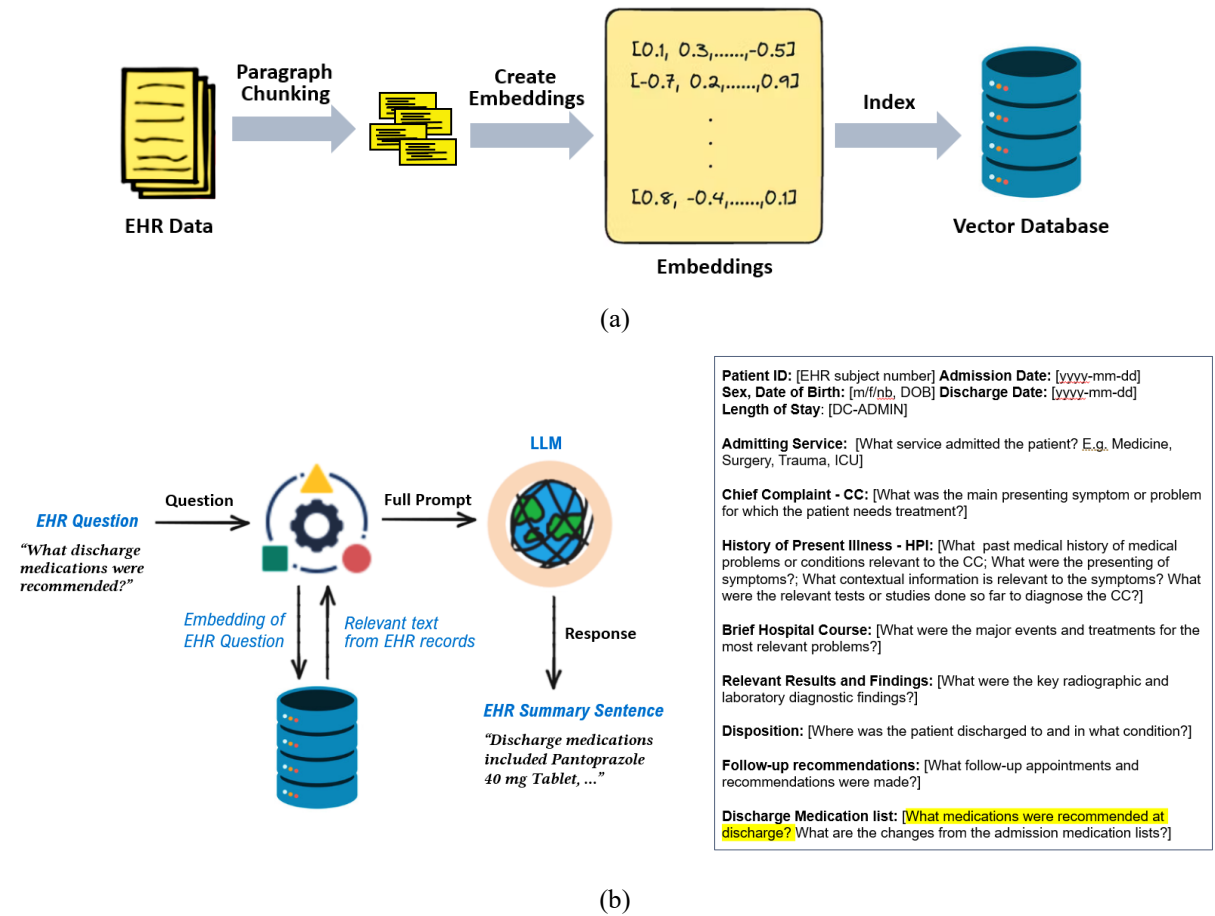

**Figure 2.** (a) EHR data is segmented into paragraphs that are indexed with their embeddings in a vector database; (b) summarization is then a process of repeated fetching of answers to the pre-defined set of questions whose answers are considered by subject-matter experts (SMEs) to form a good summary. Examples of questions that answer specific questions are shown in the right-hand figure of (b).

## 2.4 Evaluation

Summaries generated by our approach will be subjectively evaluated by subject-matter-experts for accuracy, relevance, and completeness, and objectively evaluated for accuracy, completeness, and compression ratio by the calculation of ROUGE and BLEU scores and other appropriate metrics. We will also combine the quantitative methods (ROUGE and BLEU) with semantic matching of summaries using embeddings and vector similarity.

## 3. Concluding Remarks

Summarization of electronic health records (EHRs) can save clinicians valuable screen time that currently takes one third of their time. Traditional machine learning approaches in summarization of EHRs suffer from the availability of annotated training data as well as the fact that different users may want different 'kinds of summaries' based on their role. By utilizing the latest in LLMs, semantic search and retrieval augmented generation (RAG) we can alleviate these problems by considering summarization to be the extraction of specific answers to specific questions that are deemed to be important by subject-matter experts. Our initial experiments are very promising and we are currently in the process of performing an evaluation on a large dataset. We also plan on applying the RAG/question-answering approach to summarization in other domains – for example in HR where the summarization of interviews can also be reduced to extracting specific answers to specific questions.